%% file: main.tex
\documentclass[sigconf,nonacm]{acmart}

\setcopyright{none}

\usepackage{algorithm}
\usepackage{algorithmic}
\usepackage{xspace}
\usepackage{soul}
\usepackage{subcaption}

\newcommand{\method}{NgramRL\xspace}
\sethlcolor{gray!25}
\hypersetup{pdflang={en-US}}
\AtBeginDocument{}

\begin{document}

\title[NgramRL: Rollout-Aware N-gram Speculative Decoding]{NgramRL: Accelerating RL Post-Training for Structured Generation via Rollout-Aware N-gram Speculative Decoding}

\author{Nilesh Prasad Pandey}
\orcid{0009-0002-1118-0345}
\authornote{Both authors contributed equally to this research.}
\email{nppandey@ucsd.edu}
\affiliation{%
  \institution{University of California San Diego}
  \city{La Jolla}
  \state{CA}
  \country{USA}
}

\author{Jangseon Park}
\orcid{0009-0005-3508-3859}
\email{jap036@ucsd.edu}
\authornotemark[1]
\affiliation{%
  \institution{University of California San Diego}
  \city{La Jolla}
  \state{CA}
  \country{USA}
}

\author{Jason Kong}
\orcid{0009-0000-1345-6558}
\email{jckong@ucsd.edu}
\affiliation{%
  \institution{University of California San Diego}
  \city{La Jolla}
  \state{CA}
  \country{USA}
}

\author{Onat Gungor}
\orcid{0000-0001-7215-0890}
\email{onat.gungor@mail.wvu.edu}
\affiliation{%
  \institution{West Virginia University}
  \city{Morgantown}
  \state{WV}
  \country{USA}
}

\author{Tajana Rosing}
\orcid{0000-0002-6954-997X}
\email{tajana@ucsd.edu}
\affiliation{%
  \institution{University of California San Diego}
  \city{La Jolla}
  \state{CA}
  \country{USA}
}

\renewcommand{\shortauthors}{Pandey and Park et al.}

\begin{abstract}
Reinforcement learning (RL) post-training improves the reasoning
capabilities of large language models, but its cost is dominated by
autoregressive rollout generation, which can exceed 70\% of training
time. Existing speculative-decoding methods are poorly suited to this
setting: model-based drafters consume scarce GPU memory and may become
misaligned as the policy evolves, while existing model-free methods do
not explicitly exploit repetition across training epochs. We observe
that recent rollouts of the same prompt provide a high-quality,
self-updating draft source. Based on this insight, we propose
NgramRL, a per-prompt $n$-gram drafter constructed from
bounded recent rollout history. NgramRL requires no auxiliary
model, drafter training, or persistent GPU-resident drafter state,
while target-model verification preserves the rollout distribution. To
hide CPU cache lookup and scheduling, we further introduce an adaptive
fan-out prefetcher that predicts likely verification outcomes $(j, b)$,
consisting of the accepted draft length $j$ and target-emitted
continuation token $b$. A causal budget allocator concentrates
prefetches on high-probability outcomes, while uncovered outcomes fall
back to sequential lookup without changing committed tokens. Across two
model families and three reasoning and structured-generation workloads,
\textsc{NgramRL} commits 3.60--7.77 tokens per target-model forward in
the warm-cache regime, up to $1.99\times$ more than SuffixDecoding and
$7.0\times$ more than prompt-lookup decoding. On Spider, the prefetcher
covers 96.6\% of Qwen3-8B outcomes at a candidate budget of 16 and
improves measured rollout throughput by $1.38\times$ over synchronous
NgramRL, approaching its $1.40\times$ overlap ceiling.
\end{abstract}


\maketitle

\input{section/introduction}
\input{section/related_work}
\input{section/method}
\input{section/results}
\input{section/conclusion}

\begin{acks}
This work was funded in part by NSF (awards \#1826967, \#1911095,
\#2003279, \#2052809, \#2100237, \#2112167, and \#2112665), and by
PRISM and CoCoSys, centers in JUMP 2.0, an SRC program sponsored by DARPA.
\end{acks}

\bibliographystyle{ACM-Reference-Format}
\bibliography{refs}

\end{document}

%% file: section/introduction.tex
\section{Introduction}

Reinforcement learning post-training~\cite{ouyang2022training, schulman2017proximal}, and Group Relative Policy Optimization (GRPO) in particular~\cite{shao2024deepseekmath}, has become central to improving LLM reasoning~\cite{guo2025deepseek}. But it is expensive, and the cost is dominated by the rollout generation. Each GRPO step autoregressively samples many completions per prompt before any gradient is computed, and this phase alone can exceed 70\% of step time~\cite{ji2025thinking}. Accelerating rollout is therefore the most direct route to reducing the RL post-training cost. Speculative decoding~\cite{leviathan2023fast,chen2023accelerating} offers a natural solution for improving the efficiency of RL using a small drafter that proposes tokens which the target model verifies in parallel, preserving the target's exact output distribution. However, this mechanism does not transfer cleanly to the RL rollout setting, for two reasons. First, a separate draft model or a set of trained draft heads~\cite{li2024eagle,cai2024medusa} consumes GPU memory alongside the policy, optimizer states, and KV cache, and must itself be trained and maintained, an added cost precisely where memory and compute are already scarce. Second, such drafters degrade over training. As the policy drifts, a fixed drafter's proposals are increasingly rejected~\cite{liu2023online}, so acceptance falls over exactly the long horizons on which RL runs. To overcome these issues, model-free drafters~\cite{saxena2023prompt,he2024rest} remove the additional memory cost by reusing
tokens from the prompt or a datastore rather than predicting them, but their draft candidates are confined to a single generation, due to which acceptance stays low on the adapting traces that RL post-training produces. 

Our key insight is that during RL training, the rollout already contains its own drafter, for free. Because GRPO revisits the same prompts across steps and epochs, and the policy changes only incrementally between steps, a prompt's completions are highly repetitive with those generated for it earlier in training. Prior rollouts are thus a self-updating source of draft tokens requiring no extra parameters, no training, and automatically tracking the policy since they are produced by it. This repetition is especially strong for structured tasks, whose outputs are regular by construction, and it arises across rollouts of the same prompt over training, a signal that previous model-free drafters, which draw only on a single generation, cannot exploit~\cite{saxena2023prompt,he2024rest}.

Exploiting this signal, however, raises a systems problem. A model-free cache carries no GPU memory cost precisely because it lives on the CPU, but a naive integration then serializes the CPU lookup with the GPU forward: the GPU stalls on the host every decode step while the next draft is looked up and the step is scheduled. In our measurements this host round trip idles the GPU for nearly a third of each iteration, and the cache lookup is a negligible part of it; the synchronization and scheduling dominate. Realizing the cache's benefit therefore requires co-designing the drafter with the rollout engine so that the host work is hidden behind the GPU, not merely making the lookup fast.

To this end, we propose \method{}, a rollout-aware per-prompt n-gram cache that acts as a model-free drafter for speculative decoding during RL post-training. For each prompt, \method{} maintains an
n-gram table built from that prompt's prior rollouts and proposes continuations by lookup, which the policy verifies. The cache lives in host memory and is queried on the CPU, overlapping GPU computation so that drafting adds no persistent GPU memory and negligible GPU time, and because it is populated by the policy's own outputs it stays aligned with the policy rather than going stale like a fixed drafter. To keep this CPU work off the critical path, \method{} pairs the cache with an \emph{asynchronous multipath prefetch} runtime that prefetches the next step's drafts during verification, so the decode iteration collapses to GPU busy time.

We make the following contributions:
\begin{itemize}
    \item We identify and characterize prompt-conditioned temporal locality in RL rollouts. GRPO repeatedly samples the same prompts while updating the policy incrementally, previous rollouts provide high-quality drafts for future epochs. Across consecutive epochs, a prior rollout covers 48-96\% of a later rollout, with recurring spans of 7-23 tokens.
    
    \item We propose NgramRL, a self-updating, per-prompt $n$-gram drafter constructed entirely from recent policy rollouts. Its bounded-history design limits host-memory growth and reduces mismatch with the evolving policy, while requiring no auxiliary model, drafter training, or persistent GPU-resident drafter state.

    \item We design an adaptive fan-out prefetcher, guided by a causal budget allocator, that speculatively prepares next-step drafts for likely verification outcomes $(j, b)$ during the forward. Covered outcomes bypass the verification-dependent host round trip; uncovered requests fall back to sequential lookup without altering the target-emitted token or rollout distribution.

    \item Across two model families and three workloads, NgramRL commits 3.60--7.77 tokens per target-model forward in the warm-cache regime, up to $1.99\times$ more than SuffixDecoding and $7.0\times$ more than prompt-lookup decoding. At $N = 16$, the prefetcher covers up to 99.5\% of Spider outcomes and delivers up to $1.38\times$ measured rollout speedup over synchronous NgramRL.

\end{itemize}

%% file: section/related_work.tex
\section{Background and Related Work}
\label{sec:related}

\subsection{Background: The Rollout Bottleneck}
\label{sec:background}

A GRPO~\cite{shao2024deepseekmath} training step consists of three stages. The policy samples a group of completions (rollouts) for each prompt, a reward model scores them, and the policy is updated from the group-relative advantages. The first stage, rollout, dominates the wall-clock cost. Completions are produced autoregressively, one token at a time, and this stage alone can exceed 70\% of step time~\cite{ji2025thinking}. Production frameworks such as verl~\cite{sheng2025hybridflow} therefore delegate rollout to a dedicated inference engine~\cite{kwon2023efficient, zheng2024sglang}. Speculative decoding is the natural way to accelerate it, but three properties of the RL setting make existing variants not well-suited for this problem.

\textbf{(C1)} First, GPU memory is already saturated. The policy, optimizer states, gradients, and KV cache contend for device memory during training. A speculative scheme that introduces a draft model or trained draft heads adds a persistent claim on the limited GPU resource, and can incur additional GPU-side speculation cost~\cite{chen2024sequoia}.  \textbf{(C2)} Second, the target is non-stationary. Existing speculative decoding assumes a fixed target, whereas RL updates the policy at every step. A drafter trained against one snapshot is therefore stale at the next, and its acceptance declines as the policy drifts~\cite{liu2023online}. \textbf{(C3)} Third, rollouts are free-form yet repetitive across steps. Reasoning traces rarely have much overlap with the prompt, so drafters that copy from the current context find few matches. The repetition instead lies across training steps, where the same prompts recur, and hence single-generation drafters fail to use this characteristic. NgramRL is designed to satisfy \textbf{C1--C3} at once.

\subsection{Related Work}
\label{sec:relwork}

\textbf{Model-based speculative decoding.} Speculative
decoding~\cite{leviathan2023fast,chen2023accelerating} pairs a fast drafter with the target, which verifies drafted tokens in parallel while preserving the target's output distribution. Strong methods train the drafter as a separate draft model, additional decoding heads such as Medusa~\cite{cai2024medusa}, or feature-level draft models such as EAGLE~\cite{li2024eagle}. The crucial difference between this literature and our work is that these drafters add trainable parameters and assume a fixed target. In RL post-training both assumptions fail, since the parameters consume scarce GPU memory and a fixed drafter cannot track a policy that changes every step.

\textbf{Model-free drafters.} A separate line copies previously seen tokens rather than predicting them. Prompt-lookup decoding~\cite{saxena2023prompt} matches $n$-grams against the current prompt, REST~\cite{he2024rest} retrieves continuations from a fixed datastore, and Token Recycling~\cite{luo2025turning} reuses candidates discarded during a single decode. These add no parameters, but their evidence is limited to a single generation or a static corpus, so acceptance is low on the free-form traces that RL produces. The closest work is SuffixDecoding~\cite{oliaro2026suffixdecoding}, which caches token sequences from prior outputs in CPU-side suffix trees and
performs best on structured tasks such as text-to-SQL. The crucial difference is the setting. SuffixDecoding serves a fixed model, and its repetition arises from agentic inference pipelines. We operate inside RL training, where the policy is
non-stationary and the repetition arises from GRPO revisiting the same prompts. Our per-prompt cache consequently co-evolves with the policy, and acceptance strengthens as training proceeds, a behavior the fixed-target setting does not exhibit.

\textbf{Structured generation.} The highest-value targets for RL post-training, including text-to-SQL, code generation, and tool and API calls, produce outputs with strong syntactic regularity. Completions of the same prompt share schema tokens, keywords, and boilerplate, so identical tokens recur, and copy-based drafting is known to favor such low-entropy outputs~\cite{oliaro2026suffixdecoding}. We show that this regularity is amplified in RL training, where the same prompts
are sampled repeatedly, so structured tasks benefit most from rollout-derived
$n$-gram caching.

%% file: section/method.tex
\section{NgramRL: Rollout-Aware N-gram Speculative Decoding}
\label{sec:method}

\subsection{Overview}
\label{sec:method-overview}

\begin{figure}[tbp]
    \centering
    \includegraphics[width=0.92\linewidth]{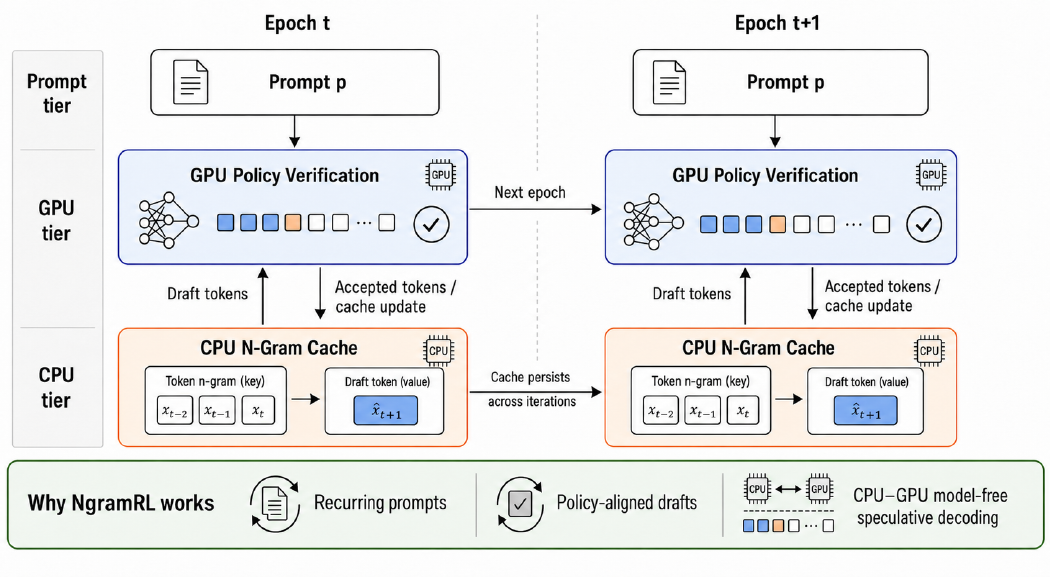}
    \caption{Overview of NgramRL's CPU-GPU model-free speculative decoding across training epochs.}
    \label{fig:framework}
\Description{Two adjacent epochs show the same prompt entering GPU policy verification. A per-prompt CPU n-gram cache proposes draft tokens and receives accepted tokens to update its history. Cache contents persist into the next epoch, connecting recurring prompts to policy-aligned, model-free drafts.}
\end{figure}

As shown in Figure~\ref{fig:framework}, NgramRL is a co-design of a model-free drafter with the RL rollout engine. It accelerates rollout without adding parameters, training a draft model, or claiming persistent GPU memory.

The design rests on one property of RL post-training. GRPO~\cite{shao2024deepseekmath} revisits the same prompts across steps and epochs, and the policy changes only incrementally between them, so a prompt's completions repeat those it produced earlier in training. Figure~\ref{fig:overlap-examples} illustrates this phenomenon: rollouts of the same Spider~\cite{yu2018spider} (text-to-SQL) prompt one epoch apart share only the query skeleton early on and become verbatim identical by late training. NgramRL turns this redundancy into accepted drafts, a signal a single-generation drafter cannot utilize.

\begin{figure}[tbp]
\centering
\begin{minipage}{\linewidth}
\footnotesize\ttfamily\raggedright
\textrm{\textbf{Early (step 10 $\to$ 115, 50\% shared):}}\\
\textrm{R1:} \hl{SELECT COUNT(*) FROM Companies WHERE Industry} IN ('Banking', 'Conglomerate');\\
\textrm{R2:} \hl{SELECT COUNT(*) FROM Companies WHERE Industry} = 'Banking' OR Industry = 'Conglomerate';\\[3pt]
\textrm{\textbf{Mid (step 300 $\to$ 407, 81\% shared):}}\\
\textrm{R1:} \hl{SELECT billing\_country, COUNT(*)} as count \hl{FROM invoices GROUP BY billing\_country ORDER BY} count \hl{DESC LIMIT 5;}\\
\textrm{R2:} \hl{SELECT billing\_country, COUNT(*)} AS num\_invoices \hl{FROM invoices GROUP BY billing\_country ORDER BY} num\_invoices \hl{DESC LIMIT 5;}\\[3pt]
\textrm{\textbf{Late (step 1200 $\to$ 1305, 100\% shared):}}\\
\textrm{R1:} \hl{SELECT b.bname FROM bank b WHERE b.state = 'New York';}\\
\textrm{R2:} \hl{SELECT b.bname FROM bank b WHERE b.state = 'New York';}
\end{minipage}
\caption{Same-prompt Spider rollouts one epoch apart (R1: earlier; R2: later).
Shared spans are highlighted; overlap increases as training progresses.}
\label{fig:overlap-examples}
\Description{Three pairs of SQL rollouts from successive epochs illustrate shared text increasing with training. The early pair shares SELECT COUNT and the table and column names but differs in its predicate. The middle pair differs mainly in its count alias. The late pair is identical. Shared spans are shaded and the reported overlap is 50, 81, and 100 percent, respectively.}
\end{figure}

Each prompt owns an $n$-gram cache built from its prior rollouts. During rollout
the cache proposes continuations by lookup and the policy verifies them exactly,
preserving the output distribution. The cache lives in host memory and is queried
on the CPU, so drafting adds no persistent GPU memory and overlaps GPU compute. The
remaining subsections detail the cache, the draft-and-verify procedure, and the
CPU-GPU co-design.

\subsection{Per-Prompt N-gram Cache}
\label{sec:method-cache}

Based on this insight, we design our proposed drafter as a set of per-prompt $n$-gram look-up tables (LUTs), denoted $\mathcal{C}_p$. We keep one table per prompt rather than a single shared table, because rollouts of the same prompt are more repetitive with each other than rollouts of different prompts. The cache is built and maintained in two steps as summarized in Algorithm~\ref{alg:cache}:

\textbf{Cache Construction:} For a prompt $p$, let $R_p^{(t)}$ be the rollouts sampled for $p$ at training step $t$. After each step, we extract $n$-grams of order $n_{\min}$ to $n_{\max}$ from these rollouts and insert them into the table $\mathcal{C}_p$, which maps each context of $n{-}1$ tokens to its observed continuation and frequency. The orders $n_{\min}$ and $n_{\max}$ are a design choice, trading memory vs drafting quality.

\textbf{Self-Updating Behavior:} $\mathcal{C}_p$ is rebuilt continually from the policy's own recent rollouts, so as the policy updates its continuations converge toward the policy's current preferences. The drafter thus tracks the policy without retraining. To bound table size and stay aligned with the latest policy, we retain rollouts from a recent window of $W$ steps and evict older entries.

\begin{algorithm}[tbp]
\caption{Incremental per-prompt cache update at step $t$}
\label{alg:cache}
\begin{algorithmic}[1]
\REQUIRE prompt $p$, cache $\mathcal{C}_p$, order range $[n_{\min}, n_{\max}]$, window $W$
\STATE $R_p^{(t)} \gets$ rollouts sampled for $p$ at step $t$
\FORALL{$n$-grams $(x_{i-n+1}, \dots, x_i)$ in $R_p^{(t)}$, \ $n \in [n_{\min}, n_{\max}]$}
  \STATE $ctx \gets (x_{i-n+1}, \dots, x_{i-1})$ \COMMENT{construction: insert / count}
  \STATE $\mathcal{C}_p[ctx][x_i].count \mathrel{+}{=} 1$; \ \ $\mathcal{C}_p[ctx][x_i].step \gets t$
\ENDFOR
\STATE \textit{Self-update:} evict $(ctx, x)$ with $step < t{-}W{+}1$
\end{algorithmic}
\end{algorithm}
\subsection{Draft-and-Verify}
\label{sec:method-draft}

Once the table is constructed, at every training step $t$ during the rollout of prompt $p$, NgramRL proposes draft tokens from $\mathcal{C}_p$ and the policy $\pi_\theta$ verifies them in one forward pass.

\textbf{Draft Proposal:} We query $\mathcal{C}_p$ for the longest matching n-gram of the current context and propose its saved continuation as the draft token. If no match exists, the position falls back to standard decoding.

\textbf{Verification Preserves the Policy:} Let $x$ be the drafted token at a position with context $c$. NgramRL accepts $x$ with probability

\begin{equation}
p_{\text{acc}}(x) = \min\!\left(1,\ \frac{\pi_\theta(x \mid c)}{q(x \mid c)}\right),
\label{eq:accept}
\end{equation}
where $q$ is the cache's proposal distribution; on rejection, it resamples from the normalized positive residual $(\pi_\theta(\cdot\mid c)-q(\cdot\mid c))_+$. This is the standard speculative-sampling correction~\cite{leviathan2023fast,chen2023accelerating}, and it guarantees that the accepted token is distributed exactly as $x \sim \pi_\theta(\cdot \mid c)$,
independent of the drafter $q$. Consequently the rollouts, rewards, advantages, and gradients are identical in distribution to non-speculative training, so NgramRL leaves accuracy and convergence unchanged.

\subsection{N-Gram Prefetcher}
\label{sec:method-prefetch}

\begin{figure*}[tp]
\centering
\includegraphics[width=0.82\textwidth]{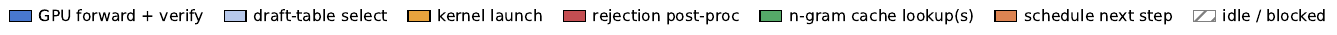}\\[2pt]
\begin{subfigure}[t]{0.49\textwidth}
  \centering
  \includegraphics[width=\linewidth]{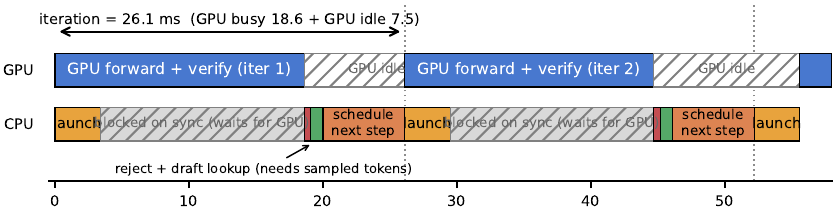}
  \caption{Synchronous baseline.}
  \label{fig:timeline-sync}
\end{subfigure}\hfill
\begin{subfigure}[t]{0.49\textwidth}
  \centering
  \includegraphics[width=\linewidth]{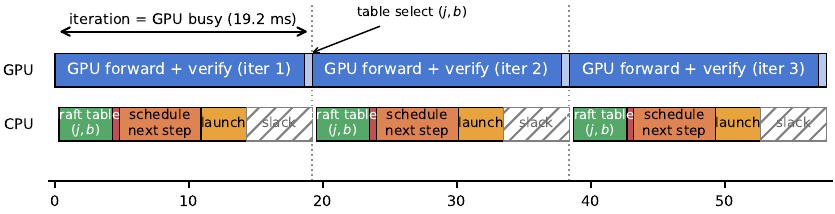}
  \caption{Asynchronous fan-out prefetching (NgramRL).}
  \label{fig:timeline-async}
\end{subfigure}
\caption{Decode-iteration timelines illustrating the mechanism. Prefetching overlaps host work with GPU verification, reducing the iteration from 26.1 to 19.2\,ms}
\label{fig:timeline}
\Description{Two CPU and GPU timelines compare synchronous lookup with asynchronous fan-out prefetching. In the synchronous case, CPU launch and scheduling wait on GPU verification, leaving exposed GPU idle intervals in a 26.1-millisecond iteration. With prefetching, draft-table construction and scheduling overlap GPU execution, and the depicted iteration takes 19.2 milliseconds.}
\end{figure*}

Placing the cache on the CPU keeps it out of GPU memory but serializes the two processors. As Figure~\ref{fig:timeline}(a) shows, each decode iteration is a host--device round trip: the GPU verifies the draft, and only once the result is synchronized to the host can the CPU look up the next draft, schedule the step, and launch its kernels, leaving the GPU idle for ${\sim}29\%$ of every iteration. The lookup is a negligible part of this gap; the synchronization and scheduling dominate, so hiding the lookup alone is not enough. \method{} therefore removes the whole round trip with a fan-out prefetcher that prepares the next step's drafts during the current forward.

\subsubsection{Adaptive Bonus-Token Fan-out Prefetcher}
Overlapping the lookup with the forward is hard because the lookup context for step $t{+}1$ is fixed only by the verification outcome of step $t$, a pair $(j, b)$: the \emph{accept length} $j \in \{0, \dots, k\}$ and the \emph{bonus token} $b$ the target emits at position $j{+}1$ (resampled on a rejection, or the free token after a full accept). The accept length takes $k{+}1$ values and is known to the CPU, but $b$ is sampled on the GPU from the full vocabulary and ends the next context, so it is unknown until verification finishes. Rather than discard $b$, \method{} \emph{predicts} it, which keeps every committed token target-distributed and preserves the lossless guarantee of Section~\ref{sec:method-draft}.

During step $t$'s forward, the prefetcher builds a draft table for step $t{+}1$ keyed by the outcome $(j, b)$. For each funded accept-length branch $j$, it draws the top-$f_j$ bonus candidates from $\mathcal{C}_p$, the frequency-ranked successors of the context extended by the $j$ accepted tokens, and issues one lookup per candidate to build the corresponding $k$-token draft; the finished table is transferred to the GPU before the forward returns. Verification then yields the realized pair $(j^{*}, b^{*})$, which the GPU resolves by an $O(1)$ key probe with no content search. On a \emph{hit}, with probability $c$ (the \emph{coverage}), the selected draft is identical to the one the sequential design would have built, and decoding proceeds with no exposed host time; on a \emph{miss}, the entry is discarded and only that request issues the ordinary sequential lookup, so misses never stall the batch. In either case the accepted prefix and the bonus token are committed unchanged, so \method{} stays lossless (Eq.~\eqref{eq:accept} holds) and degrades at worst to the synchronous baseline.

\textbf{Adaptive budget allocation.} The candidate budget $N{=}\sum_j f_j$ is split across branches from trailing statistics over a causal window of $W$ steps: the accept-length histogram $\hat{P}(j)$ and each branch's coverage profile $\hat{c}_j(f)$, the probability that the true bonus ranks within the top $f$ candidates. We maximize expected coverage,
\begin{equation}
\max_{\{f_j\}} \sum_j \hat{P}(j)\,\hat{c}_j(f_j) \quad \text{s.t.} \quad \sum_j f_j = N,
\label{eq:alloc}
\end{equation}
solved greedily by marginal gain. In effect the fan-out concentrates around the recent mode of the accept-length distribution, and full-accept branches, whose bonus is the next token of an already-cached trajectory and is the top candidate ${\sim}90\%$ of the time, need only one candidate ($f_j{=}1$), so a small budget covers high-acceptance tasks.

\textbf{Cost.} Let $T_{\text{gpu}}$ be the device-busy time per iteration and $T_{\text{cpu}}$ the serial host time it must hide. A covered iteration hides the host chain and a missed one pays it, giving
\begin{equation}
\text{speedup}(c) = \frac{T_{\text{gpu}}+T_{\text{cpu}}}{T_{\text{gpu}}+(1-c)\,T_{\text{cpu}}} \;\xrightarrow{\,c\to1\,}\; 1 + \frac{T_{\text{cpu}}}{T_{\text{gpu}}}.
\label{eq:speedup}
\end{equation}
The prefetch chain itself costs ${\approx}N$ lookups per drafted request and must fit under $T_{\text{gpu}}$; beyond $N{\approx}64$ it approaches the host time $T_{\text{cpu}}$ it hides, so the budget sweet spot is small (Section~\ref{sec:results-prefetch}). Both the attainable speedup and the sweet spot are governed by the ratio $T_{\text{cpu}}/T_{\text{gpu}}$: larger models or larger batches raise $T_{\text{gpu}}$ relative to the host chain, so the hidden overlap is a smaller fraction of each iteration, whereas CPU-heavier regimes benefit more.

\subsection{Integration and Generality}
\label{sec:method-integration}

We integrate NgramRL into verl's~\cite{sheng2025hybridflow} vLLM rollout worker. The cache attaches to the rollout engine and needs no change to the reward model, optimizer, or policy architecture, so it is a drop-in optimization. Adding no parameters and no training, it composes with existing rollout-efficiency methods and applies to any RL-style setup where prompts recur.

%% file: section/results.tex
\section{Results}
\label{sec:results}

\subsection{Experimental Setup}

\subsubsection{RL Training Setup} For all our experiments, we use
verl~\cite{sheng2025hybridflow} for GRPO training~\cite{shao2024deepseekmath},
running on $2{\times}$ NVIDIA A100 80GB GPUs. The rollouts are served by
the vLLM inference engine~\cite{kwon2023efficient}, sampling 5 completions for each of 64 prompts
per step over 15 epochs. We adopt verl's default GRPO hyperparameters for
each model and use the same settings across all baselines. Speculative
decoding is exact, so no drafter alters the training dynamics, the
sampled rollouts, or the resulting policy. The comparison therefore
isolates drafting efficiency alone, and we evaluate all drafters on
identical training runs.

\textbf{Models:} We evaluate on Qwen3-8B~\cite{yang2025qwen3} and
DeepSeek-LLM-7B~\cite{bi2024deepseek}, two popular open-weight
models from different families, to demonstrate generalization.

\textbf{Datasets:} For our evaluation, we use GSM8K~\cite{cobbe2021training},
Spider~1.0~\cite{yu2018spider}, and CodeContests~\cite{li2022competition},
chosen to span free-form and structured generation. GSM8K is free-form
mathematical reasoning. Spider (text-to-SQL) and CodeContests (code)
produce syntactically regular outputs, where cross-rollout repetition is
strongest. This spread lets us characterize how acceptance scales with
the regularity of the output.

\textbf{Baselines:} We compare against three model-free
drafters under a matched draft budget of $k{=}16$. Prompt-lookup decoding
(PLD)~\cite{saxena2023prompt} drafts $n$-gram continuations from the
current context only. REST~\cite{he2024rest} retrieves continuations from
a fixed datastore. SuffixDecoding~\cite{oliaro2026suffixdecoding}
retrieves from a global suffix tree over the most recent 10k responses.

\textbf{Metrics:} We report tokens per decoding step (TPS): the tokens
committed per target forward pass, i.e.\ one plus the mean accepted draft
length. It is the standard lossless-speedup proxy in speculative decoding
and is hardware-agnostic, so the comparison does not depend on a specific
GPU or kernel. We report two aggregates: the mean over the full training run, and the mean over
the final 30\% of steps. We use the final 30\% because training has
largely saturated by this point and the cache is fully populated, which
is where speculative decoding helps the most. We also plot TPS across the
entire run, showing how it evolves as the policy trains.

\subsubsection{Profiling Setup}
Runtime measurements use a dedicated rig: Qwen3-8B (bf16, batch 64, $k{=}16$) on one NVIDIA RTX PRO 6000 (96\,GB), with a 32-core AMD Ryzen Threadripper 9970X host holding the $n$-gram cache and running the lookup/scheduling chain. The throughput results (Figure~\ref{fig:speedup}) are measured on this 8B rig. Per task, a timing pass records iteration wall-clock and hit/accept statistics, and an NVIDIA Nsight Systems~\cite{nsight} pass decomposes each iteration period into device-busy time (the union of kernel, memory-copy, and CUDA-graph intervals) and exposed host time; serial host work (launch, rejection handling, cache lookup, scheduling) that are not overlapped with device execution and therefore stall the GPU. 
\begin{figure*}[!t]
    \centering
    \includegraphics[width=0.86\textwidth]{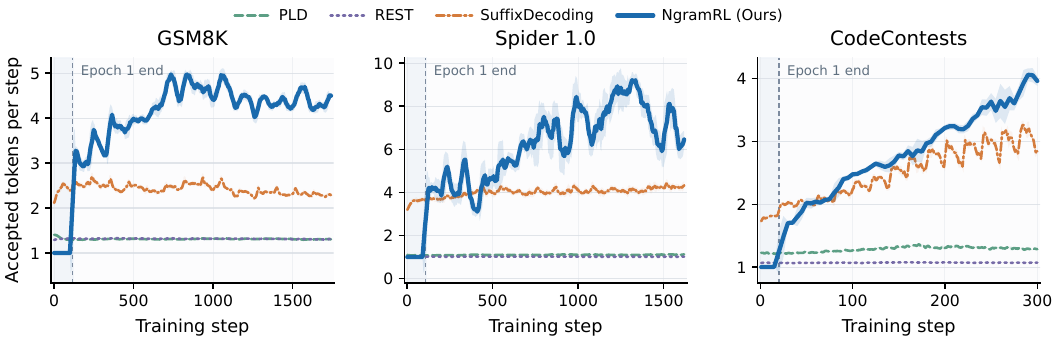}
    \caption{Tokens per decoding step over GRPO training with Qwen3-8B on GSM8K,
    Spider~1.0, and CodeContests. The dashed line marks the end of the first epoch.}
    \label{fig:tps-dynamics}
\Description{Three line charts compare PLD, REST, SuffixDecoding, and NgramRL over training steps on GSM8K, Spider 1.0, and CodeContests. NgramRL starts near one token per step while the cache is cold, then increases after prompts recur. Its largest gains occur on Spider. PLD is dashed, REST dotted, SuffixDecoding dash-dot, and NgramRL solid, so the methods can also be distinguished without color. The end of the first epoch is marked by a vertical dashed line.}
\end{figure*}

\begin{table}[tbp]
\centering
\caption{Tokens committed per decoding step (TPS; $k{=}16$), as overall training mean /
mean over the final 30\% (the stable regime; see Metrics). Best per task in
\textbf{bold}.}
\label{tab:main}
\small
\setlength{\tabcolsep}{3pt}
\begin{tabular}{@{}lccc@{}}
\toprule
Method & GSM8K & Spider & CodeCont. \\
\midrule
\multicolumn{4}{l}{\textit{Qwen3-8B}} \\
PLD~\cite{saxena2023prompt}                & 1.31 / 1.31 & 1.10 / 1.11 & 1.28 / 1.30 \\
REST~\cite{he2024rest}                     & 1.32 / 1.31 & 1.02 / 1.02 & 1.07 / 1.07 \\
SuffixDec.~\cite{oliaro2026suffixdecoding} & 2.42 / 2.32 & 4.01 / 4.15 & 2.50 / 2.96 \\
NgramRL                                       & \textbf{4.00 / 4.36} & \textbf{5.94 / 7.77} & \textbf{2.70 / 3.60} \\
\midrule
\multicolumn{4}{l}{\textit{DeepSeek-7B}} \\
PLD~\cite{saxena2023prompt}                & 1.47 / 1.49 & 1.24 / 1.24 & 1.19 / 1.15 \\
REST~\cite{he2024rest}                     & 1.69 / 1.69 & 1.02 / 1.02 & 1.02 / 1.02 \\
SuffixDec.~\cite{oliaro2026suffixdecoding} & 3.37 / 3.64 & 3.71 / 3.88 & 3.11 / 3.99 \\
NgramRL                                       & \textbf{5.83 / 6.81} & \textbf{6.37 / 7.71} & \textbf{3.92 / 6.19} \\
\bottomrule
\end{tabular}
\end{table}

\subsubsection{Accepted-Length Comparison and Dynamics}
Table~\ref{tab:main} reports acceptance as the full-run mean and the final-30\% mean. In the stable regime our cache reaches 3.6 to 7.8 tokens per step. This is a $1.2$ to $1.99\times$ improvement over SuffixDecoding~\cite{oliaro2026suffixdecoding} and up to $7.0\times$ over PLD~\cite{saxena2023prompt}. The overall means include our cold first epoch and
still lead on every task. The advantage tracks output regularity. It is largest where completions are prompt-specific, as on Spider and DeepSeek's GSM8K derivations. It is smallest on CodeContests, where cross-prompt boilerplate favors
a global cache.

\begin{figure}[tbp]
    \centering
    \includegraphics[width=0.80\linewidth]{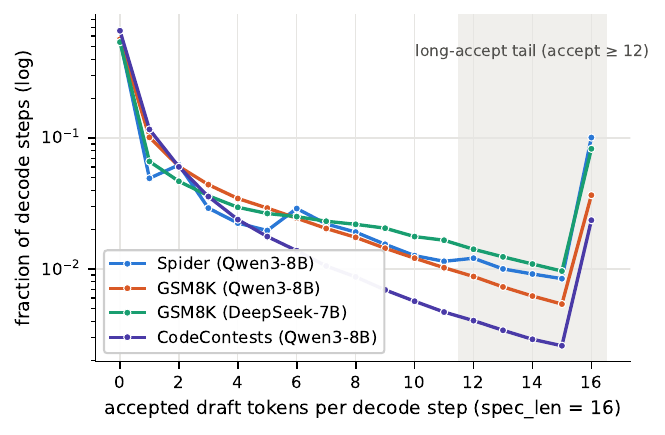}
    \caption{Accepted-length distribution using $k{=}16$, aggregated over each run.}
    \label{fig:accept-dist}
\Description{A log-scale line chart shows the fraction of decode steps at each accepted draft length from zero through sixteen for four model and task combinations. Most mass occurs at zero, while all curves retain a long-acceptance tail and rise at full acceptance of sixteen draft tokens.}
\end{figure}

\textit{Heavy-tailed acceptance:} A key benefit of our cache is that it captures long verbatim replays that other drafters miss. The mean accepted length hides a skewed distribution. Figure~\ref{fig:accept-dist} shows the accepted length
is heavy-tailed. On many steps the policy accepts most or all of the 16 draft tokens. On Spider, 12 or more are accepted on $14.1\%$ of steps and all 16 on up to $10.1\%$ ($13.0\%$, $6.4\%$, and $3.7\%$ of long-accept steps on DeepSeek GSM8K,
Qwen GSM8K, and CodeContests). These are verbatim replays of a prior epoch's solution, verified in one pass. A small draft budget would forfeit them, so our large budget of $k{=}16$ is what turns this tail into rollout speedup.

Next, we examine how acceptance evolves as training proceeds. Figure~\ref{fig:tps-dynamics} shows the central dynamic on all three tasks. Acceptance is zero until prompts recur, then increases as history accumulates. On GSM8K and CodeContests acceptance strengthens throughout training. On Spider it peaks mid-run and then declines, as the policy's SQL outputs diversify and older rollouts go stale, an effect we quantify in the window ablation below. On CodeContests the global cache leads in early training, while per-prompt history is least effective, and our cache overtakes both the baselines once that history accumulates. The redundancy is direct to measure. Consecutive-epoch rollouts of the same Spider prompt share substrings of 7 to 23 tokens, and a cache of only the earlier rollout already covers 48 to $96\%$ of the later one.

\begin{figure}[tbp]
  \centering
  \begin{subfigure}[t]{0.49\linewidth}
    \centering
    \includegraphics[width=\linewidth]{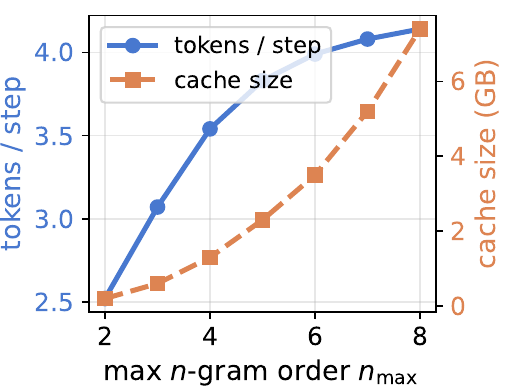}
    \caption{TPS and memory vs.\ $n_{\max}$.}
    \label{fig:nmax}
  \end{subfigure}\hfill
  \begin{subfigure}[t]{0.49\linewidth}
    \centering
    \includegraphics[width=\linewidth]{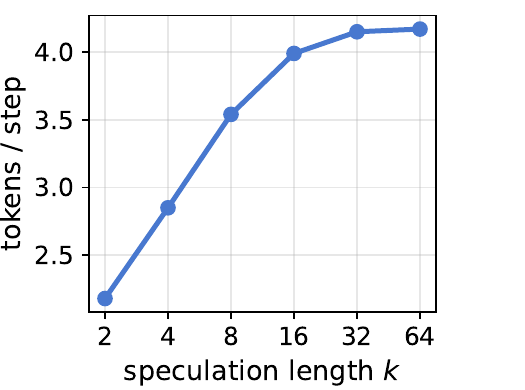}
    \caption{TPS vs.\ speculation length $k$.}
    \label{fig:speclen}
  \end{subfigure}
  \caption{Hyperparameter ablations (GSM8K, Qwen3-8B). We fix $n_{\max}{=}6$ and
  $k{=}16$, each at the knee of its speedup curve.}
  \label{fig:ablations}
\Description{Two plots show Qwen3-8B on GSM8K. Increasing maximum n-gram order raises tokens per step with diminishing gains beyond order six, while host cache memory rises rapidly. Increasing speculation length raises tokens per step with diminishing gains beyond sixteen tokens.}
\end{figure}

\subsubsection{N-gram Cache Design Ablations}
Four design choices govern the cache: two drafting hyperparameters, the maximum
$n$-gram order $n_{\max}$ and the speculation length $k$, and two structural
choices, per-prompt keying and a bounded history window. The hyperparameters
trade draft quality against cache memory, while the structural choices determine
what the cache indexes and retains, and therefore where its gains come from.
We ablate each in turn.

\textbf{Maximum $n$-gram order:} As shown in Figure~\ref{fig:ablations}
for Qwen3-8B on GSM8K, a higher $n_{\max}$ raises acceptance but grows the cache
super-linearly, and acceptance saturates beyond order 6, so we set $n_{\max}{=}6$.

\textbf{Speculation length:} A longer $k$ likewise raises acceptance
until it saturates near 16, and leaves the cache unchanged, so we set $k{=}16$.
Both values sit at the knee of their speedup curves, and although
Figure~\ref{fig:ablations} shows one model and dataset, we find they transfer
well to the others, so we use the same configuration for all experiments in
the paper.

\textbf{Per-prompt keying:} With the drafting budget fixed, we turn to
the cache's structure. Table~\ref{tab:keying} shows the gain comes from prompt
keying, not the data structure. Replacing our per-prompt keys with a single
global bucket drops acceptance to $2.41$, close to SuffixDecoding at $2.43$,
while per-prompt keying reaches $3.05$ over the same steps. This gap is not
caused by a small baseline cache. Increasing SuffixDecoding's cache fourfold,
to 40k responses, raises it only to $2.65$. The limit is the lack of
prompt-conditioned lookup, not cache size.

\textbf{History window:} Table~\ref{tab:window} sweeps the per-prompt
window $w$, the number of recent appearances the cache retains. A window of 4
captures 79 to $91\%$ of the full acceptance at a fraction of the memory. Beyond
$w{=}16$, more history does not help and slightly lowers acceptance on GSM8K, as
older rollouts go stale. This is the same drift behind Spider's mid-run decline
in Figure~\ref{fig:tps-dynamics}. A small window therefore saves memory and
keeps the cache aligned with the current policy.

\begin{table}[tbp]
\centering
\caption{Global vs per-prompt cache design ablation.}
\label{tab:keying}
\footnotesize
\begin{tabular}{lcc}
\toprule
Configuration & Cold epoch & Plateau \\
\midrule
Ours, global bucket       & 1.55 & 2.41 \\
Ours, per-prompt keys     & 1.00 & \textbf{3.05} \\
SuffixDecoding~\cite{oliaro2026suffixdecoding} & 1.90 & 2.43 \\
\bottomrule
\end{tabular}
\end{table}

\begin{table}[tbp]
\centering
\caption{History-window ablation: accepted tokens per decode step (cache MB) as the
per-prompt window $w$ grows.}
\label{tab:window}
\small
\setlength{\tabcolsep}{3pt}
\begin{tabular}{@{}lcccc@{}}
\toprule
 & \multicolumn{4}{c}{Accepted / step (cache MB)} \\
\cmidrule(lr){2-5}
Task & $w{=}1$ & $w{=}4$ & $w{=}16$ & $w{=}\infty$ \\
\midrule
\multicolumn{5}{l}{\textit{Qwen3-8B}} \\
GSM8K    & 1.87 (841) & 2.81 (1878) & 3.38 (3181) & 3.34 (3634) \\
Spider   & 3.39 (236) & 5.13 (255)  & 5.82 (309)  & 5.82 (309) \\
CodeCont.& 1.72 (249) & 2.64 (591)  & 2.89 (1742) & 2.91 (1744) \\
\midrule
\multicolumn{5}{l}{\textit{DeepSeek-7B}} \\
GSM8K    & 3.69 (177) & 5.08 (293)  & 5.74 (662)  & 5.74 (662) \\
Spider   & 3.00 (230) & 4.63 (251)  & 5.83 (335)  & 5.83 (335) \\
CodeCont.& 3.73 (67)  & 5.12 (89)   & 5.78 (290)  & 5.78 (293) \\
\bottomrule
\end{tabular}
\end{table}

\subsubsection{N-gram Prefetcher}
\label{sec:results-prefetch}
\begin{figure}[tp]
\centering
\begin{subfigure}[b]{0.56\linewidth}
  \centering
  \includegraphics[height=2.9cm]{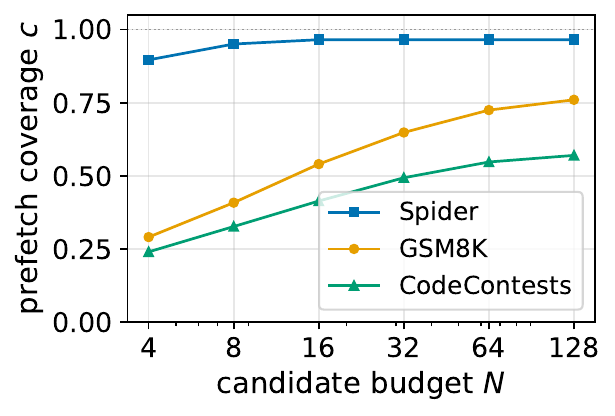}
  \caption{Coverage $c$ vs.\ budget $N$.}
  \label{fig:coverage}
\end{subfigure}\hfill
\begin{subfigure}[b]{0.42\linewidth}
  \centering
  \includegraphics[height=2.9cm]{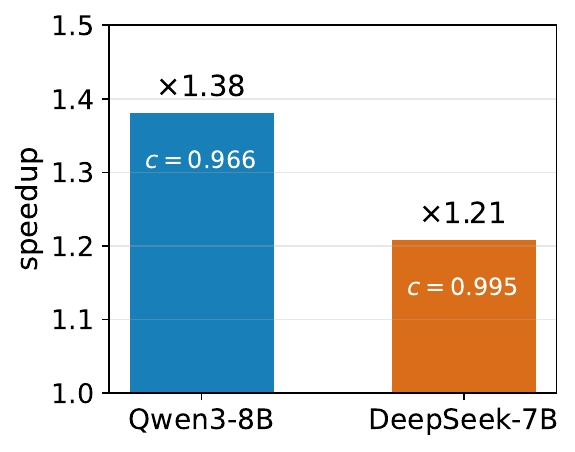}
  \caption{Speedup at $N{=}16$.}
  \label{fig:speedup}
\end{subfigure}
\caption{Fan-out prefetcher: (a)~prefetch coverage $c$ vs.\ candidate budget $N$; (b)~Spider speedup over the sequential synchronous baseline at $N{=}16$.}
\label{fig:fanout}
\Description{The left plot shows prefetch coverage rising with candidate budget. Spider approaches full coverage with a small budget; GSM8K and CodeContests remain lower. The right plot reports Spider speedups at budget sixteen: 1.38 times for Qwen3-8B with coverage 0.966, and 1.21 times for DeepSeek-7B with coverage 0.995.}
\end{figure}

\textbf{Prefetch coverage:}
Figure~\ref{fig:coverage} reports the prefetch coverage $c$, the probability that the realized outcome $(j^{*}, b^{*})$ is among the prefetched candidates, as a function of the candidate budget $N$, measured by exact replay of the saturated Qwen3-8B cache with the causal trailing-window allocator. A small budget suffices: Spider reaches $c{=}0.95$ at $N{=}8$ and saturates at $0.97$ by $N{=}16$, while GSM8K and CodeContests climb to $0.65$ and $0.49$ at $N{=}32$ with only logarithmic gains beyond. Coverage follows the same agreement-length law as acceptance: after a full accept the bonus token is the next token of an already-cached trajectory, so high-acceptance tasks are also high-coverage tasks. The residual miss mass is entropy-limited rather than data-limited: per-step coverage plateaus within a few training steps and accumulating more rollout history does not raise it. Pushing $N$ past the knee therefore buys little while the prefetch-chain cost grows linearly, so we operate at $N{=}8$--$32$.

% Activate ACM's column balancing while still in the first column.
\balance
\textbf{End-to-end speedup:}
Figure~\ref{fig:speedup} translates coverage into wall-clock gain over the sequential synchronous baseline via Eq.~\eqref{eq:speedup}, with $T_{\text{gpu}}$ and $T_{\text{cpu}}$ the CUPTI-measured device-busy and exposed-host times per decode iteration. On Spider at $N{=}16$, Qwen3-8B achieves $1.38\times$ of its $1.40\times$ overlap ceiling ($c{=}0.966$), and DeepSeek-7B, whose ceiling we measure directly at $1.21\times$, achieves it outright ($c{=}0.995$): near-perfect coverage makes the prefetcher equivalent to removing the host chain from the critical path, with no accepted token lost and every committed token still drawn from the target distribution. On the higher-entropy tasks the same mechanism yields $1.05$--$1.06\times$ against ceilings of $1.10\times$ (GSM8K) and $1.14\times$ (CodeContests); because misses fall back to the sequential lookup per request, the worst case is the baseline itself.

%% file: section/conclusion.tex
\section{Conclusion}
We presented \textsc{NgramRL}, a rollout-aware speculative-decoding
system that exploits prompt-conditioned temporal locality in RL
post-training. \textsc{NgramRL} builds bounded per-prompt $n$-gram
caches from recent policy rollouts, yielding self-updating drafts
without an auxiliary model, drafter training, or GPU-resident drafter
state. An adaptive fan-out prefetcher with a causal budget allocator
keeps the host-resident cache off the decode critical path by preparing
drafts for likely verification outcomes $(j, b)$, while uncovered
requests fall back to sequential lookup without altering the rollout
distribution. Across two model families and three workloads,
\textsc{NgramRL} commits 3.60--7.77 tokens per target-model forward, up
to $1.99\times$ more than SuffixDecoding and $7.0\times$ more than
prompt-lookup decoding, and the prefetcher delivers up to
$1.38\times$ measured rollout speedup over synchronous execution.
These results show that recent same-prompt rollouts are an effective
draft source and that outcome-aware CPU--GPU prefetching converts this
temporal locality into measured acceleration while preserving exact
target sampling.